\pdfoutput=1

\documentclass[11pt]{article}

\usepackage[preprint]{acl}

\usepackage{times}
\usepackage{latexsym}
\usepackage{subcaption}
\usepackage{graphicx}
\usepackage{stmaryrd}
\usepackage{amsmath} 
\usepackage{amsfonts}
\usepackage[T1]{fontenc}

\usepackage[utf8]{inputenc}

\usepackage{microtype}

\usepackage{inconsolata}
\usepackage{enumitem}

\usepackage{graphicx}
\usepackage{url}

\title{An Exploration of Knowledge Editing for Arabic}

\author{
Basel Mousi \hspace{11mm} Nadir Durrani  \hspace{11mm} Fahim Dalvi \\
Qatar Computing Research Institute, HBKU, Doha, Qatar \\ 
{\tt \{bmousi,ndurrani,faimaduddin\}@hbku.edu.qa} \\ 
}

\begin{document}
\maketitle


\begin{abstract}

 While Knowledge Editing (KE) has been widely explored in English, its behavior in morphologically rich languages like Arabic remains underexamined. In this work, we present the first study of Arabic KE. We evaluate four methods (ROME, MEMIT, ICE, and LTE) on Arabic translations of the ZsRE and Counterfact benchmarks, analyzing both multilingual and cross-lingual settings. Our experiments on Llama-2-7B-chat show that parameter-based methods struggle with cross-lingual generalization, while instruction-tuned methods perform more robustly. We extend Learning-To-Edit (LTE) to a multilingual setting and show that joint Arabic-English training improves both editability and transfer. We release Arabic KE benchmarks and multilingual training for LTE data to support future research.
\end{abstract}

\section{Introduction}

Despite their impressive capabilities, LLMs suffer from a fundamental limitation: \textbf{their knowledge is static and cannot be easily updated without costly retraining or model re-deployment.} This becomes particularly problematic when models must adapt to new facts or correct outdated or incorrect information. To address this, the field of \textit{Knowledge Editing (KE)} has emerged, offering techniques to surgically modify specific factual content within an LLM without retraining from scratch \cite{wang2024knowledgeeditinglargelanguage,yao-etal-2023-editing}.

Recently, multilingual knowledge editing has garnered some attention \cite{mela-etal-2024-mass,si2024mpnleveragingmultilingualpatch,zhang-etal-2025-multilingual,wu2025editonceupdateeverywhere,LIME,durranimousidalvi}. However, the progress on Arabic remains notably limited. \textbf{Arabic NLP} poses unique challenges due to diglossia, rich morphology, and the lack of curated resources \cite{arabicnlp-2024-arabic,Guellil_2021,arabicnlp-ws-2023-arabicnlp}. The absence of Arabic-specific knowledge editing benchmarks and evaluations creates a significant barrier to understanding how existing KE methods perform in this context.

Furthermore, in today’s multilingual world, updating knowledge in one language should ideally generalize to others. This raises critical questions around \textit{multilingual and cross-lingual knowledge editing}: i) Can an edit made in Arabic propagate cross-lingually? ii) Do the same methods perform equally across languages? iii) How can models be trained to edit themselves effectively in multiple languages?

In this work, we present the first study of knowledge editing in Arabic. We benchmark four methods (ROME, MEMIT, ICE, and LTE) on Arabic translations of the ZsRE and Counterfact datasets, evaluating their performance in both multilingual and crosslingual settings.

A central contribution of our work is \textbf{extending the Learning to Edit (LTE) framework} to support Arabic and joint Arabic and English training. This multilingual extension improves both editability and crosslingual generalization, demonstrating that instruction-tuned models can adapt edits across languages. We find that parameter-based methods perform inconsistently across languages and exhibit poor transfer. In contrast, LTE delivers strong performance in both Arabic and crosslingual scenarios. To support future research, we release our datasets and multilingual LTE training resources.

\begin{figure}
     \centering
      \begin{subfigure}[b]{\textwidth}
         \includegraphics[width=0.45\textwidth]{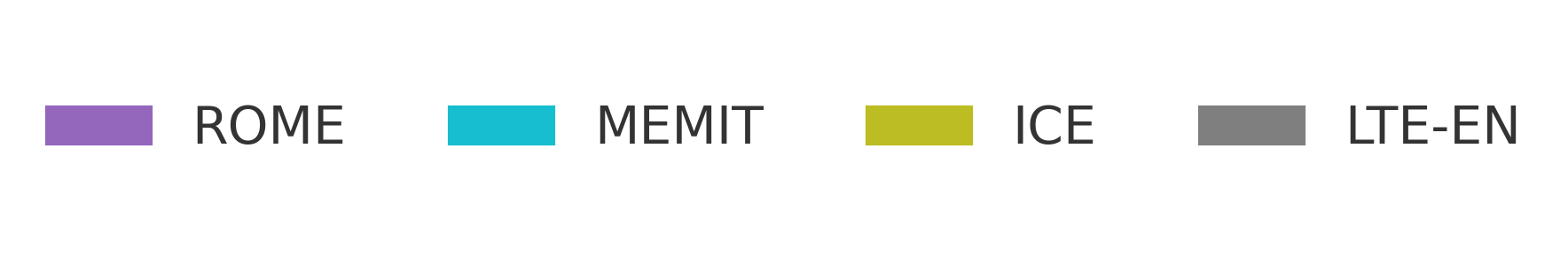}
     \end{subfigure}
     \begin{subfigure}[b]{0.45\textwidth}
         \centering
         \includegraphics[width=\textwidth]{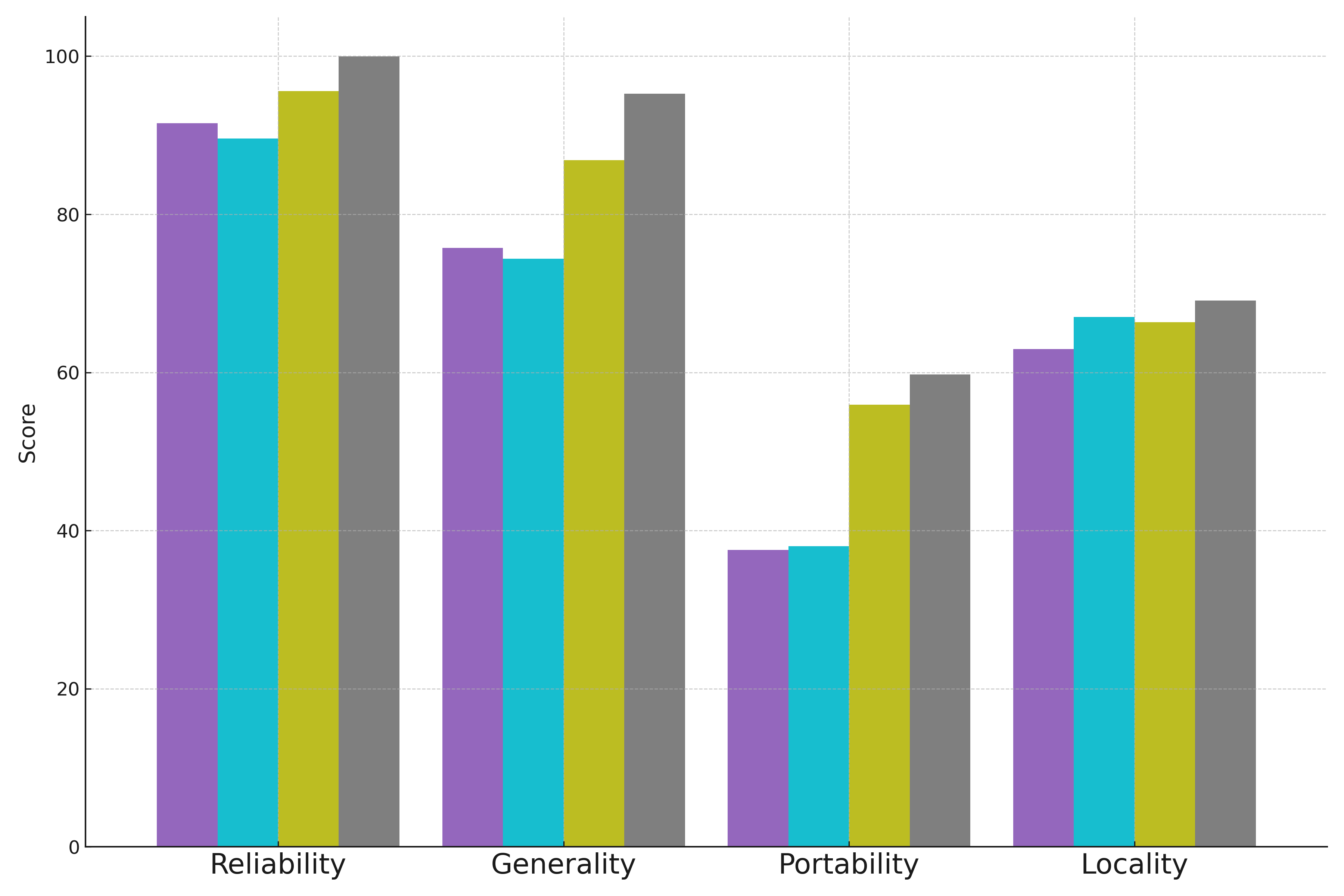}
         \label{fig:arabic_zsre_results}
     \end{subfigure}

    \caption{Comparison of ROME, MEMIT, and ICE on LLaMA2-7B-Chat across four metrics: reliability, generality, locality, and Portability}
   
    \label{fig:arabic_methods_comparison}
\end{figure}

\paragraph{Our contributions:}
\begin{itemize}[noitemsep, topsep=0pt]
    \item We analyze four KE methods (ROME, MEMIT, ICE, and LTE) on Arabic edits.
    \item We compare editing effectiveness across Arabic, English, and German.
    \item We extend LTE to multilingual settings and evaluate its crosslingual impact.
    \item We release Arabic versions of ZsRE and Counterfact for KE evaluation.
    \item We provide multilingual training data for instruction tuned editing.
\end{itemize}

\section{Preliminaries}

\textbf{Knowledge Editing (KE)} updates a language model $f_\theta$ with a new fact $(x_e, y_e)$, producing an edited model $f_{\theta_e}$ that satisfies $f_{\theta_e}(x_e) = y_e$ while preserving unrelated outputs.

We evaluate KE using four standard metrics: \textbf{reliability} (accuracy on the edit), \textbf{generality} (consistency on paraphrases), \textbf{locality} (preservation of unrelated knowledge), and \textbf{portability} (reasoning with the edited fact in new contexts).

In the \textbf{multilingual setting}, edits and evaluations occur within the same language $\ell$. In the \textbf{cross-lingual setting}, edits are applied in one language $\ell_i$ and evaluated in another $\ell_k$.

\begin{figure*}
     \centering
      \begin{subfigure}[b]{\textwidth}
         \centering
         \includegraphics[width=0.7\textwidth]{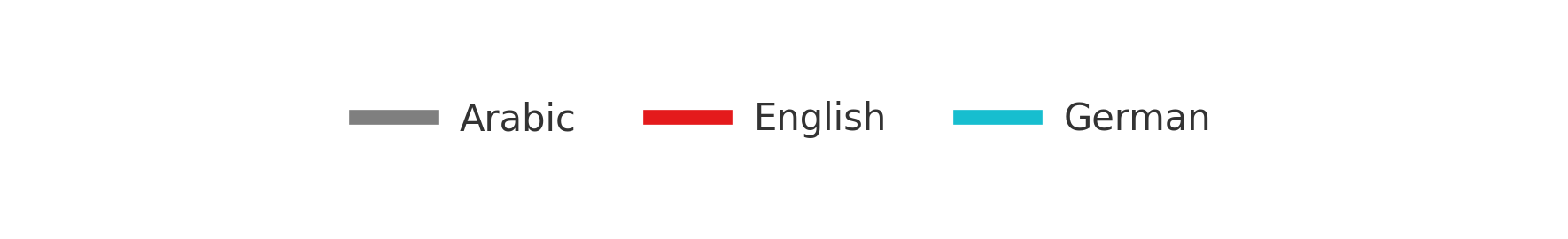}
     \end{subfigure}
     
     \begin{subfigure}[b]{0.24\textwidth}
         \centering
         \includegraphics[width=\textwidth]{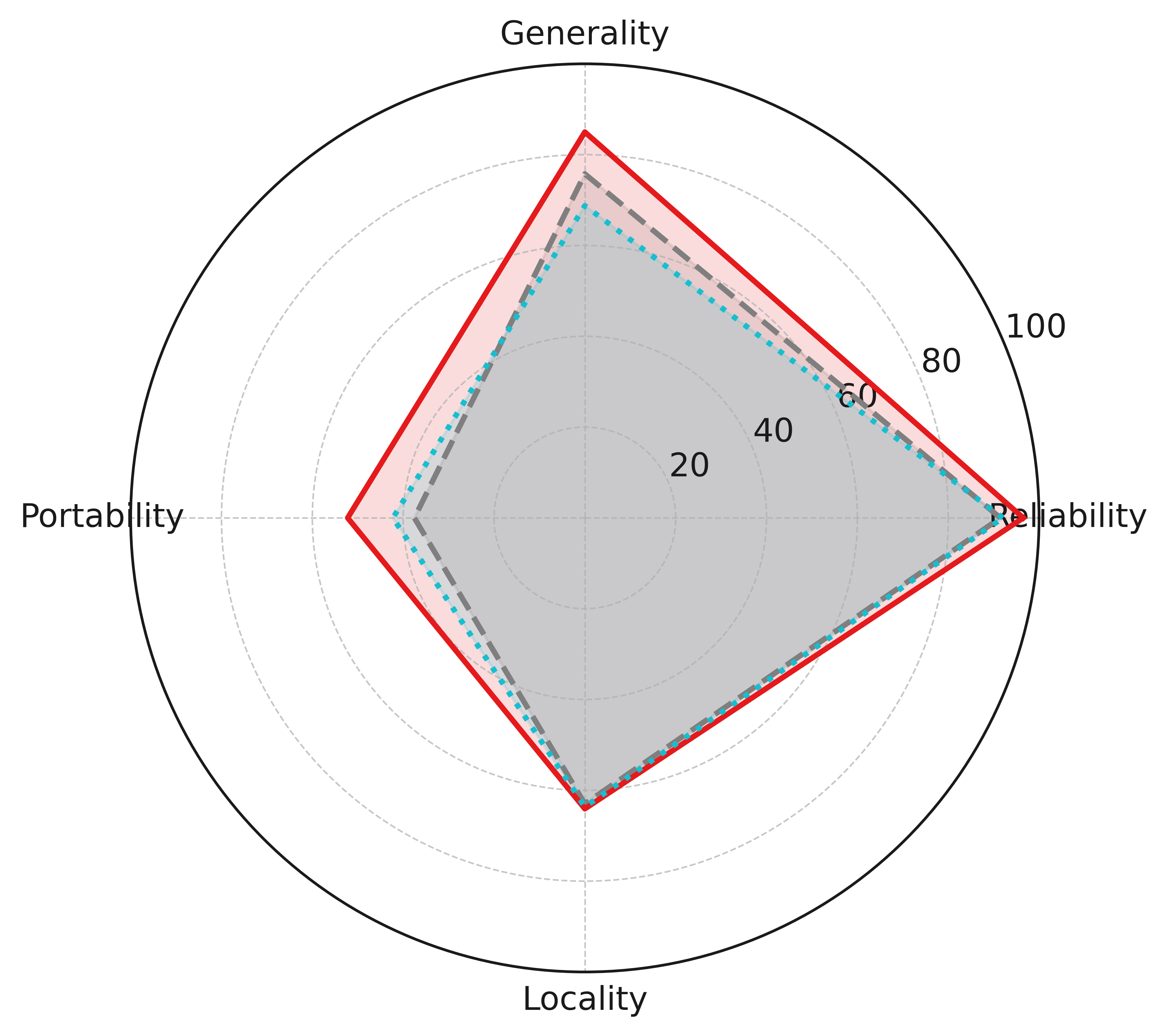}
         \caption{ROME - ZsRE}
         \label{fig:rome_languages_comparison_zsre}
     \end{subfigure}
     \begin{subfigure}[b]{0.24\textwidth}
         \centering
         \includegraphics[width=\textwidth]{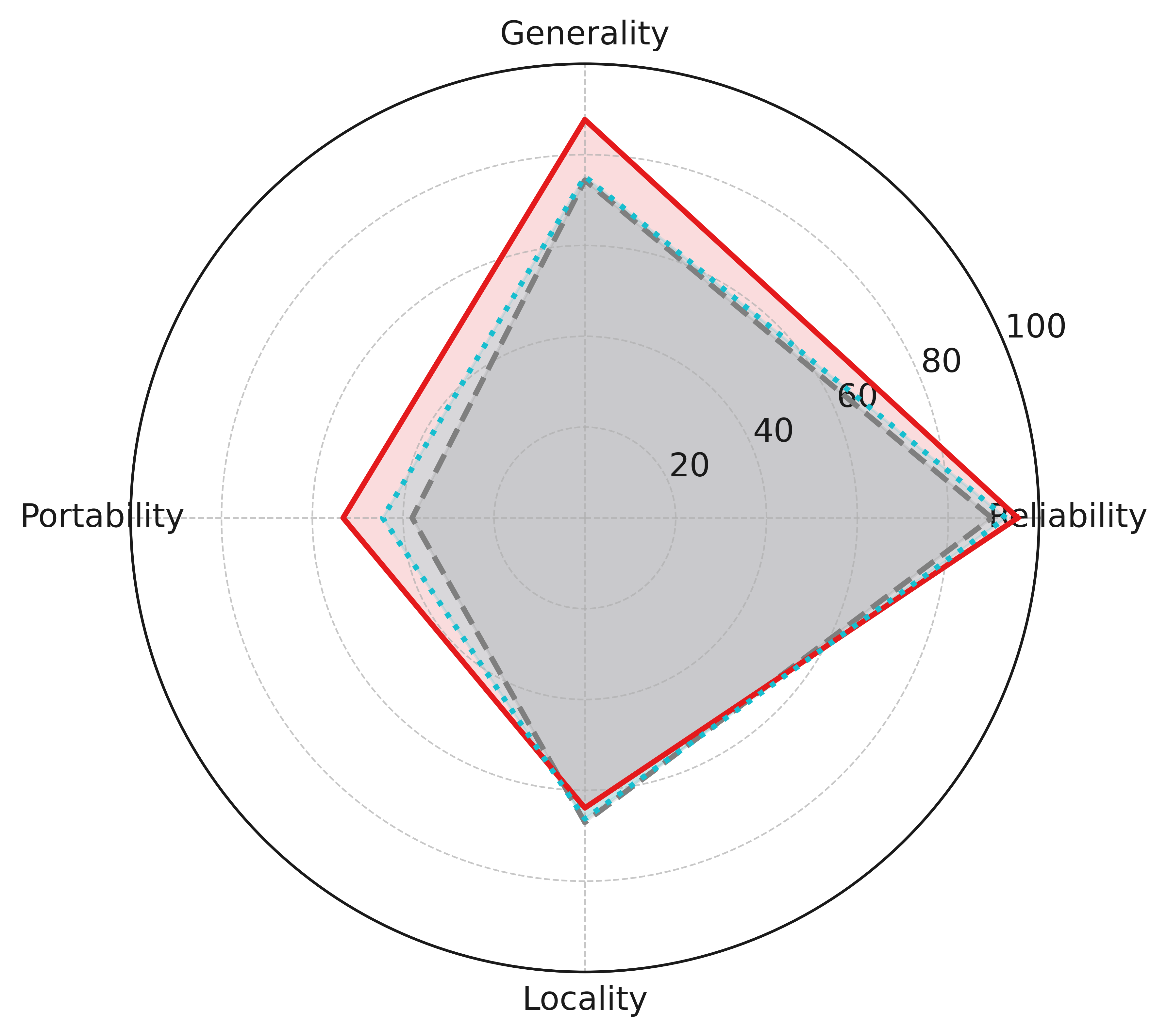}
         \caption{MEMIT - ZsRE}
         \label{fig:memit_languages_comparison_zsre}
     \end{subfigure}
      \begin{subfigure}[b]{0.24\textwidth}
         \centering
         \includegraphics[width=\textwidth]{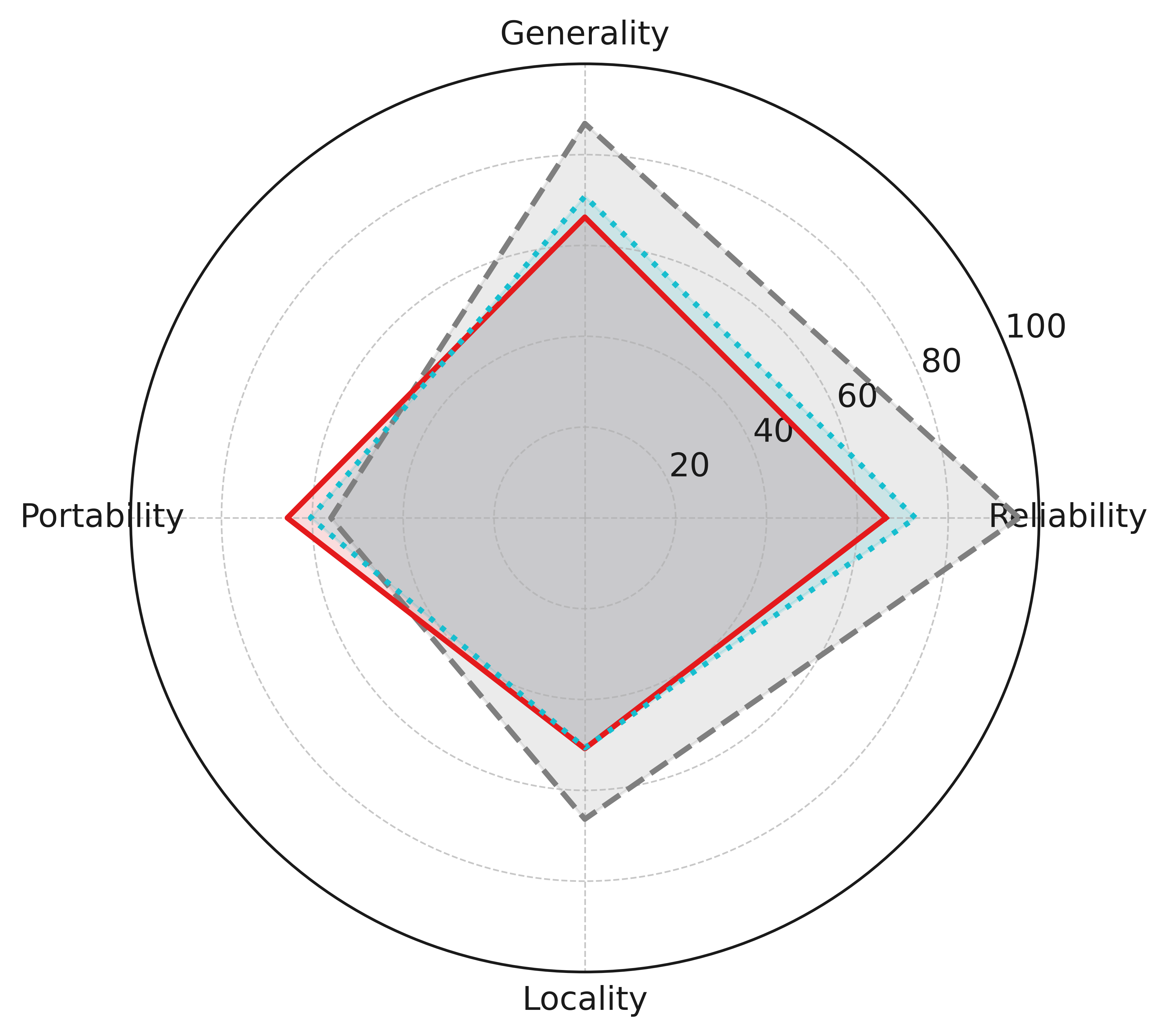}
         \caption{ICE - ZsRE}
         \label{fig:ike_languages_comparison_zsre}
     \end{subfigure}
     \begin{subfigure}[b]{0.24\textwidth}
         \centering
         \includegraphics[width=\textwidth]{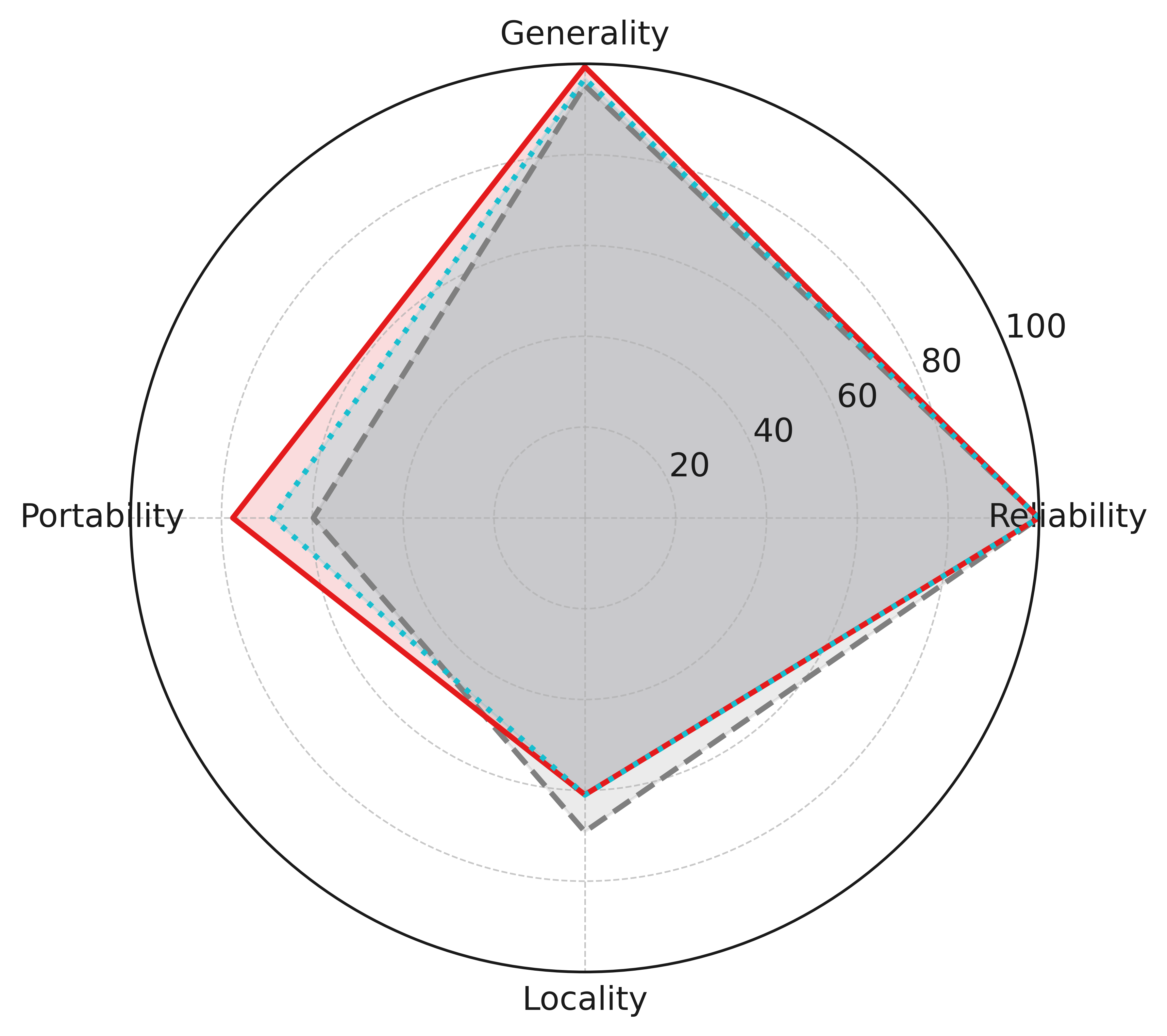}
         \caption{LTE-EN - ZsRE}
     \end{subfigure}

    \caption{Impact of the editing language on the reliability, generality, portability and locality metrics on the ZsRE and Counterfact datasets for Llama2-7B-Chat}
    \label{fig:editing_across_languages_comparison}
\end{figure*}

\section{Experimental Setup}
\subsection{Data Curation}
To enable knowledge editing research in underrepresented languages, we construct \textbf{Arabic and German versions of two widely used KE benchmarks}: \textit{ZsRE} and \textit{Counterfact}.
\paragraph{ZsRE}~\cite{levy-etal-2017-zero} was originally introduced for zero-shot relation extraction and later adapted for KE by~\cite{de-cao-etal-2021-editing, mitchell2022memorybasedmodeleditingscale}. It consists of well-defined factual triples and serves as a strong basis for evaluating \textit{reliability} and \textit{generality} in KE.
\paragraph{Counterfact}~\cite{meng2022locating} was designed to test model robustness under \textit{counterfactual} knowledge-false facts that plausibly contradict known information. This benchmark is especially useful for evaluating \textit{locality}, i.e., ensuring that edits do not bleed into unrelated knowledge.
\paragraph{Translation and Release.} We use the NLLB-200 model\footnote{\url{https://hf.co/facebook/nllb-200-3.3B}} \cite{nllbteam2022languageleftbehindscaling} to automatically translate ZsRE and Counterfact into Arabic and German. While synthetic, these translations are high-quality and provide the first large-scale KE benchmark for Arabic.  \footnote{\url{https://github.com/baselmousi/arabic-knowledge-editing}}
\paragraph{Our Contribution.} Several datasets were developed for multingual knowledge editing \cite{wei-etal-2025-mlake,wang-etal-2024-retrieval,wang-etal-2024-cross,wu2023evakellmnewbenchmarkevaluating,nie2025bmike53investigatingcrosslingualknowledge,ali-etal-2025-mqa}. To the best of our knowledge, this is the \textbf{first release of Arabic knowledge editing benchmarks} based on ZsRE and Counterfact. Each sample is aligned with evaluation protocols for \textit{reliability}, \textit{generality}, \textit{locality}, and \textit{portability}, making the data immediately usable for reproducible multilingual KE research.

We use the standardized splits from the KnowEdit benchmark~\cite{zhang2024comprehensivestudyknowledgeediting} and preserve their structure to ensure compatibility with prior work.
\subsection{Knowledge Editing Methods}

To evaluate knowledge editing in Arabic and cross-lingual contexts, we compare four representative methods spanning distinct paradigms: \textbf{ROME}~\cite{meng2022locating} and \textbf{MEMIT}~\cite{meng2022memit} (parameter-based), \textbf{ICE}~\cite{zheng-etal-2023-edit} (in-context), and \textbf{LTE}~\cite{jiang-etal-2024-learning} (instruction-tuning). While the first three offer complementary approaches to editing and generalization, our primary focus is on extending LTE, given its flexibility and potential for multilingual adaptation.

Originally designed for English, \textbf{LTE} fine-tunes models to follow edit instructions through supervised examples, enabling edits to be applied on-the-fly via prompting. We build on this framework by developing both monolingual (Arabic-only) and bilingual (Arabic+English) variants, aiming to assess how instruction diversity impacts editability in Arabic and the model’s ability to generalize across languages. This extension allows us to investigate whether LLMs can learn to edit themselves across linguistic boundaries, highlighting the promise of LTE as a foundation for scalable, instruction-driven multilingual editing.

\section{Results and Analysis}

\subsection{Arabic Editing Performance}

\paragraph{How effective are existing knowledge editing methods when applied to Arabic?} 






Figure~\ref{fig:arabic_methods_comparison} compares four editing methods: \textit{ROME}, \textit{MEMIT}, \textit{ICE}, and \textit{LTE-EN} on Arabic edits using ZsRE dataset (Counterfact results are shown in figure \ref{fig:arabic_counterfact_results} in Appendix \ref{sec:additional_results_arabic_editing}). \textbf{LTE-EN} consistently achieves the highest scores across reliability, generality, locality, and portability, indicating that instruction-tuned models, even when trained only on English, can generalize effectively to Arabic. \textbf{ICE} ranks second in reliability and generality, though its portability drops sharply on Counterfact, likely due to the challenge of counterfactual reasoning under zero-shot prompts. \textbf{MEMIT} excels in locality, preserving unrelated knowledge via its surgical update mechanism, but trails in generality and portability. \textbf{ROME} performs worst overall, highlighting the difficulty of transferring localized parameter edits to morphologically rich, non-English languages.

\subsection{Multilingual Comparison}

LLMs encode different languages in partially overlapping latent spaces \cite{mousi-etal-2024-exploring}. This raises an important research question: \textbf{How does editing in Arabic compare to editing in other languages?}



To assess cross-lingual robustness, we compare editing performance in \textbf{Arabic, English, and German} across four methods: \textit{ROME}, \textit{MEMIT}, \textit{ICE}, and \textit{LTE-EN} as shown in Figure~\ref{fig:editing_across_languages_comparison} (Counterfact results are shown in figure  \ref{fig:multilingual_comparison_counterfact}). \textbf{Parameter-based methods} (\textit{ROME} and \textit{MEMIT}) perform best in English but degrade noticeably in German and further in Arabic, reflecting their limited adaptability beyond English-tuned settings. In contrast, \textbf{ICE} exhibits stable performance across all three languages (Figure~\ref{fig:ike_languages_comparison_zsre}), suggesting that prompt-based approaches are more resilient to linguistic variation. Similarly, \textbf{LTE} shows minimal degradation across languages, highlighting the benefits of instruction tuning for multilingual generalization.
\begin{figure*}
     \centering
      \begin{subfigure}[b]{\textwidth}
         \centering
         \includegraphics[width=0.7\textwidth]{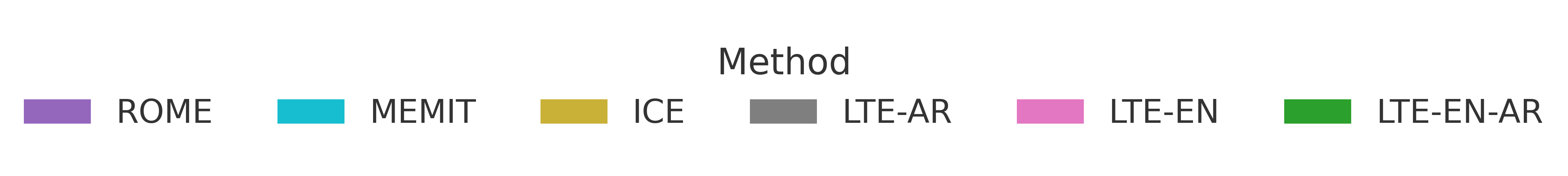}
     \end{subfigure}
     
     \begin{subfigure}[b]{0.32\textwidth}
         \centering
         \includegraphics[width=\textwidth]{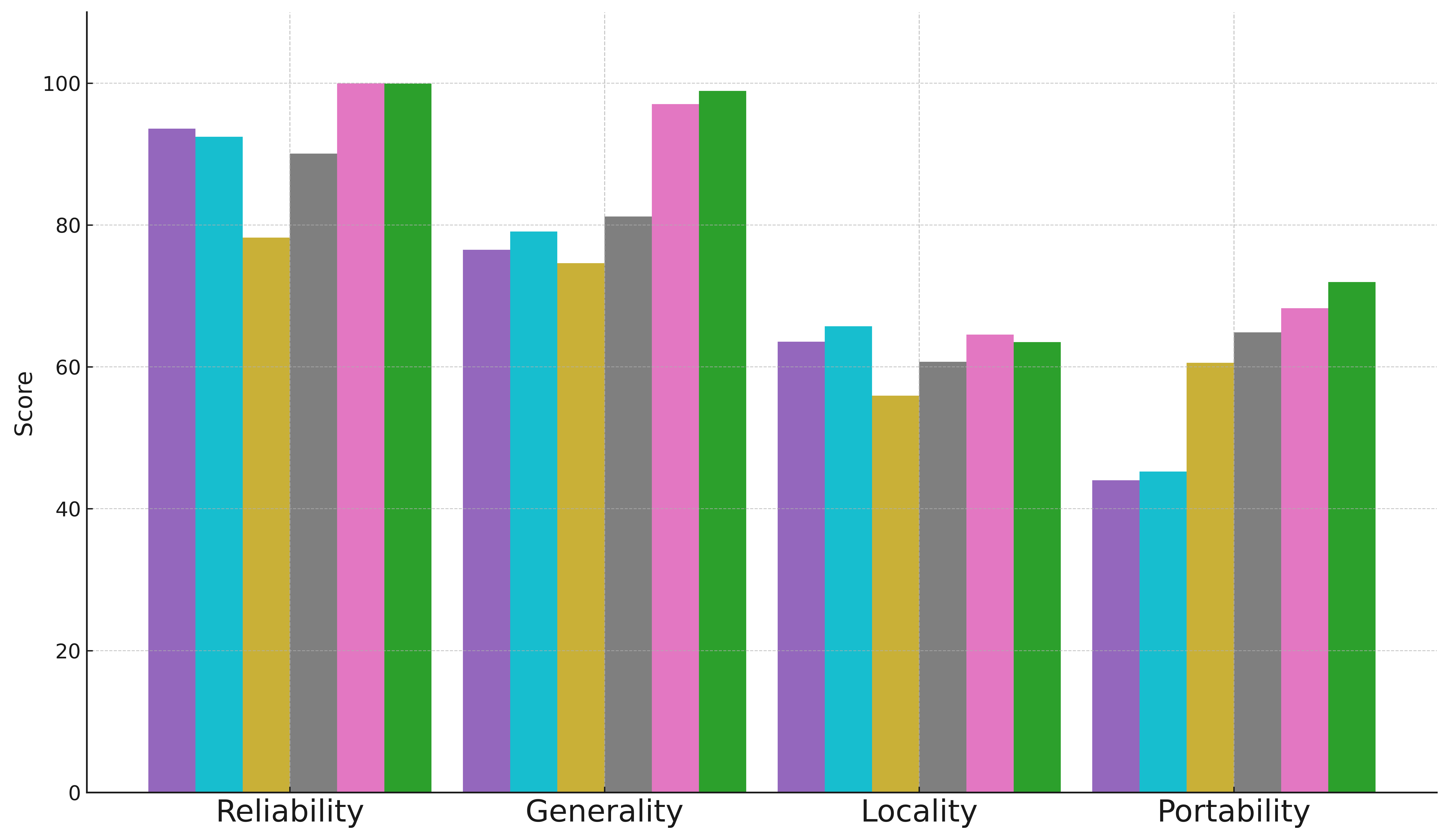}
         \caption{Multilingual Performance of Methods}
         \label{fig:multilingual-accuracy-lte}
     \end{subfigure}
     \hfill
     \begin{subfigure}[b]{0.32\textwidth}
         \centering
         \includegraphics[width=\textwidth]{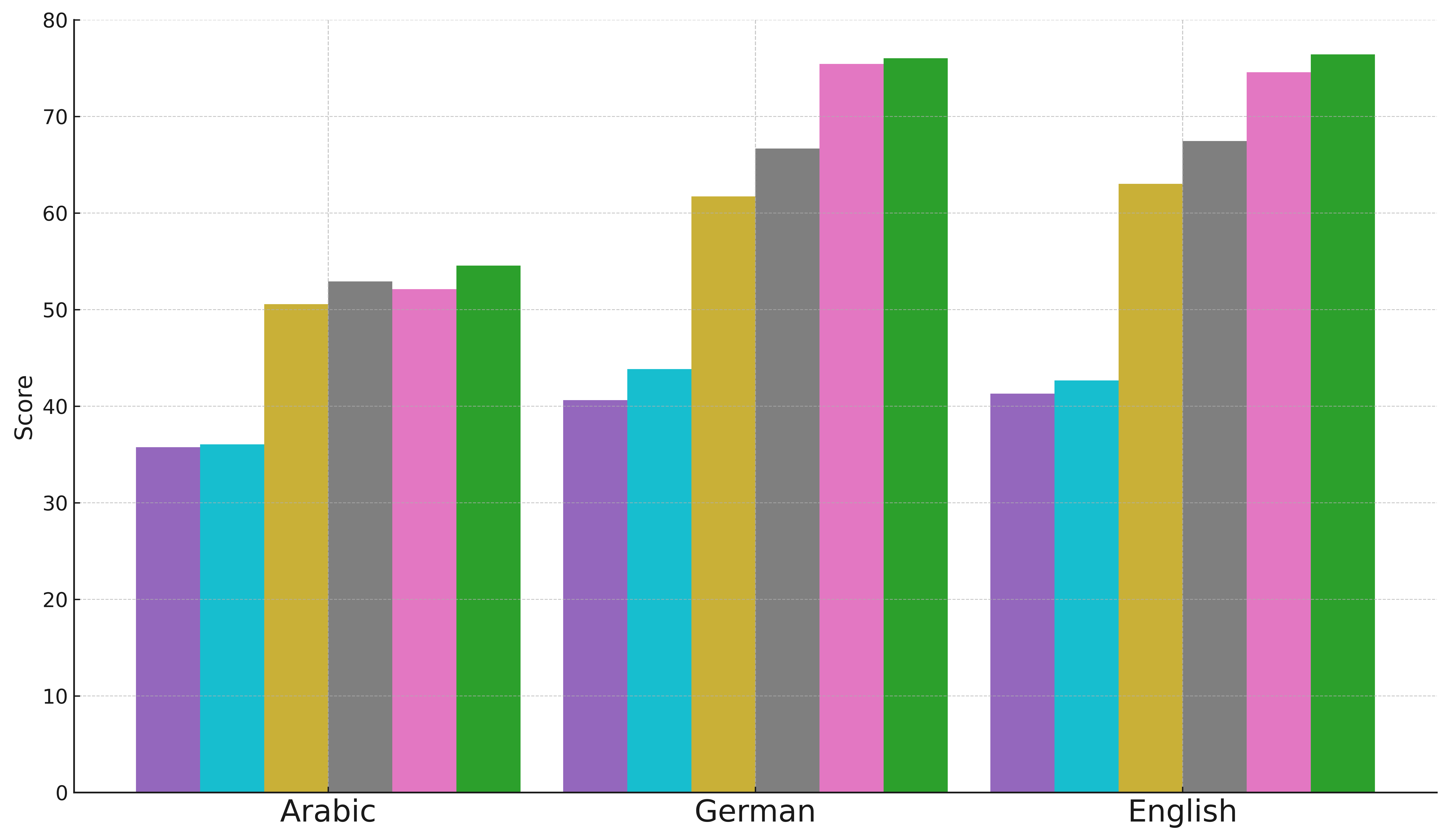}
         \caption{Cross-Lingual Reliability}
         \label{fig:crosslingual-reliability-lte}
     \end{subfigure}
     \hfill
     \begin{subfigure}[b]{0.32\textwidth}
         \centering
         \includegraphics[width=\textwidth]{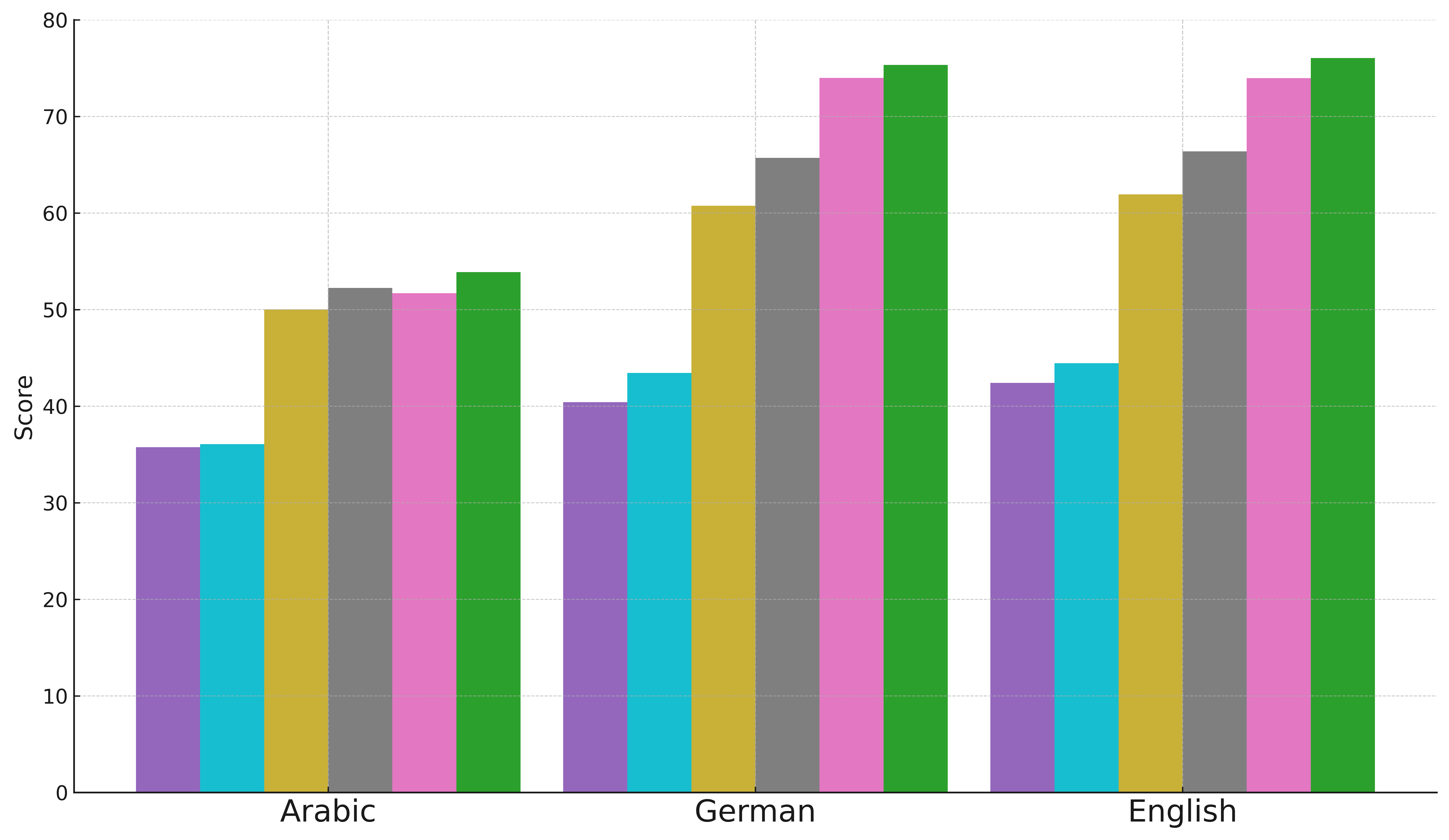}
         \caption{Cross-Lingual Generality}
         \label{fig:cross-lingual-lte-generality}
     \end{subfigure}
 
    \caption{(a) Shows a comparison of the considered methods across the \textit{reliability, generality, locality, and portability} metrics on the ZsRE dataset. (b) Shows a comparison of the averaged cross-lingual reliability scores on the ZsRE dataset and (c) Shows a comparison of the averaged cross-lingual generality scores on the ZsRE dataset. The x-axis in (b) and (c) refer to the language the edit is being applied in.}
    \label{fig:lte_metrics}
\end{figure*}
\begin{figure*}
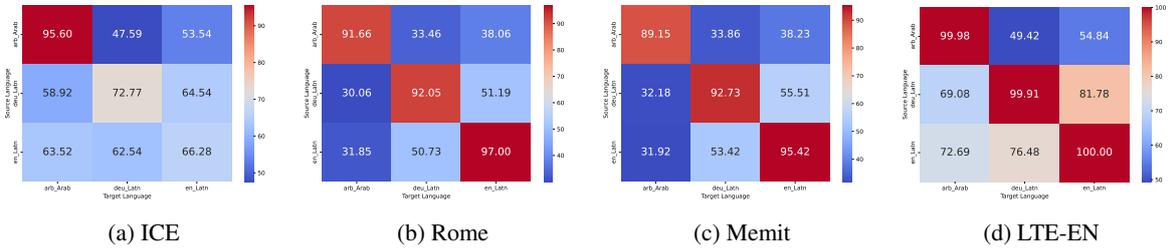

     \centering
     \begin{subfigure}[b]{0.24\textwidth}
         \centering
         \includegraphics[width=\textwidth]{figures/crosslingual-heatmaps-zsre/base_heatmap_edit_success.png}
         \caption{ICE}
         \label{fig:ice_reliability_zsre}
     \end{subfigure}
     \begin{subfigure}[b]{0.24\textwidth}
         \centering
         \includegraphics[width=\textwidth]{figures/crosslingual-heatmaps-zsre/rome-edit_success.png}
         \caption{Rome}
         \label{fig:rome_reliability_zsre}
     \end{subfigure}
     \begin{subfigure}[b]{0.24\textwidth}
         \centering
         \includegraphics[width=\textwidth]{figures/crosslingual-heatmaps-zsre/memit-edit_success.png}
         \caption{Memit}
         \label{fig:memit_reliability_zsre}
     \end{subfigure}
      \begin{subfigure}[b]{0.24\textwidth}
         \centering
         \includegraphics[width=\textwidth]{figures/crosslingual-heatmaps-zsre/english-tuned-final_heatmap_edit_success.png}
         \caption{LTE-EN}
         \label{fig:lte_reliability_zsre}
     \end{subfigure}
    \caption{Cross Lingual Reliability Metrics Comparison (ZsRE)}
    \label{fig:crosslingual_comparison_reliability_zsre}
\end{figure*}
\subsection{Cross-Lingual Transfer and Anisotropy}
\paragraph{Does editing a fact in Arabic propagate effectively to other languages, and vice versa?}
A core objective of multilingual knowledge editing is enabling factual edits to transfer seamlessly across languages~\cite{wang-etal-2024-cross,khandelwal-etal-2024-cross,beniwal-etal-2024-cross}. To test this, we evaluate bidirectional transfer performance between Arabic, English, and German using the ZsRE benchmark. We consider two setups: (a) editing in Arabic and evaluating in other languages and (b) editing in English or German and evaluating in Arabic. Figure~\ref{fig:crosslingual_comparison_reliability_zsre} reports the reliability metric on the ZsRE dataset (Appendix \ref{sec:additional_results_arabic_editing} contains additional results on the counterfact dataset). \textbf{We observe a clear asymmetry in cross-lingual transfer:} edits made in Arabic fail to propagate reliably to English or German, and vice versa. Parameter-based methods such as ROME and MEMIT show especially weak transfer, confirming that their internal representations are language-sensitive and fail to support consistent multilingual alignment. Even when editing semantically equivalent facts across languages, the models do not generalize edits effectively without explicit multilingual support.


\vspace{0.5em}
\subsection{Multilingual Learning to Edit}

\paragraph{Do instruction-tuned models generalize Arabic edits cross-lingually}

The Learning to Edit (LTE) framework~\cite{jiang-etal-2024-learning} was originally proposed to teach English models to incorporate edits through instruction tuning. We extend this framework to support Arabic and multilingual training, evaluating three variants: \textbf{LTE-EN}: Trained only on English edits, \textbf{LTE-AR}: Trained only on Arabic edits, \textbf{LTE-AR-EN}: Jointly trained on Arabic and English edits. We assess both \textit{multilingual performance} (editing and evaluating in the same language) and \textit{cross-lingual performance} (editing in one language, evaluating in another).

Figure~\ref{fig:multilingual-accuracy-lte} compares all methods across reliability, generality, locality, and portability. LTE-AR-EN outperforms all others, showing that joint multilingual training yields the most consistent and robust edit behavior. While LTE-EN performs well in Arabic despite never seeing Arabic edits, adding Arabic fine-tuning further improves generality and reliability. Notably, there is a slight drop in locality for the jointly trained model, reflecting a common trade-off between generalization and specificity.

Figures~\ref{fig:crosslingual-reliability-lte} and~\ref{fig:cross-lingual-lte-generality} further show that LTE fine-tuning substantially improves cross-lingual performance across all metrics, with LTE-AR-EN again achieving the strongest results.

\section{Conclusion}
We presented the first study of knowledge editing for Arabic, evaluating four editing paradigms: ROME, MEMIT, ICE, and LTE, on the ZsRE and Counterfact benchmarks. Our experiments reveal that parameter-based editing methods, though effective in English, struggle in Arabic and show poor crosslingual transfer. In contrast, instruction-tuned methods, especially our extended multilingual LTE framework, exhibit robust performance both in Arabic and across languages. Our findings highlight key challenges and opportunities in multilingual knowledge editing. First, language-specific morphological and syntactic factors significantly affect the reliability and locality of edits. Second, crosslingual propagation is limited in most existing approaches, emphasizing the need for multilingual training. Finally, instruction tuning emerges as a promising direction for building language-agnostic editing capabilities. We hope this work serves as a foundation for future efforts aimed at scalable and reliable knowledge editing for low-resource and morphologically rich languages like Arabic.



\newpage
\bibliography{acl_latex}
\newpage
\appendix

\section{Additional Results}
\label{sec:additional_results_arabic_editing}

\begin{figure}
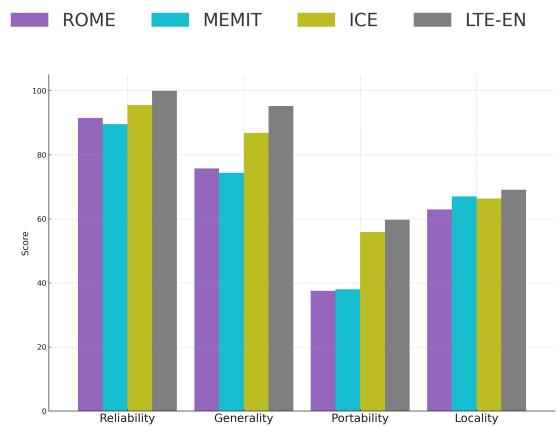

     \centering
      \begin{subfigure}[b]{\textwidth}
         \includegraphics[width=0.45\textwidth]{figures/arabic-editing/performance_metrics_legend.png}
     \end{subfigure}
     \begin{subfigure}[b]{0.45\textwidth}
         \centering
         \includegraphics[width=\textwidth]{figures/arabic-editing/performance_metrics_wider_bars_no_legend.png}
     \end{subfigure}

    \caption{Comparison of ROME, MEMIT, and ICE on LLaMA2-7B-Chat across four metrics: reliability, generality, locality, and Portability on the counterfact dataset}
    \label{fig:arabic_counterfact_results}
\end{figure}

\paragraph{Arabic Editing} The results of Arabic editing performance on the counterfact dataset are shown in figure \ref{fig:arabic_counterfact_results}. 

\begin{figure*}
     \centering
      \begin{subfigure}[b]{\textwidth}
         \centering
         \includegraphics[width=0.7\textwidth]{figures/multilingual-comparison-mod/radar_legend_arabic_english_german.png}
     \end{subfigure}
     
     \begin{subfigure}[b]{0.24\textwidth}
         \centering
         \includegraphics[width=\textwidth]{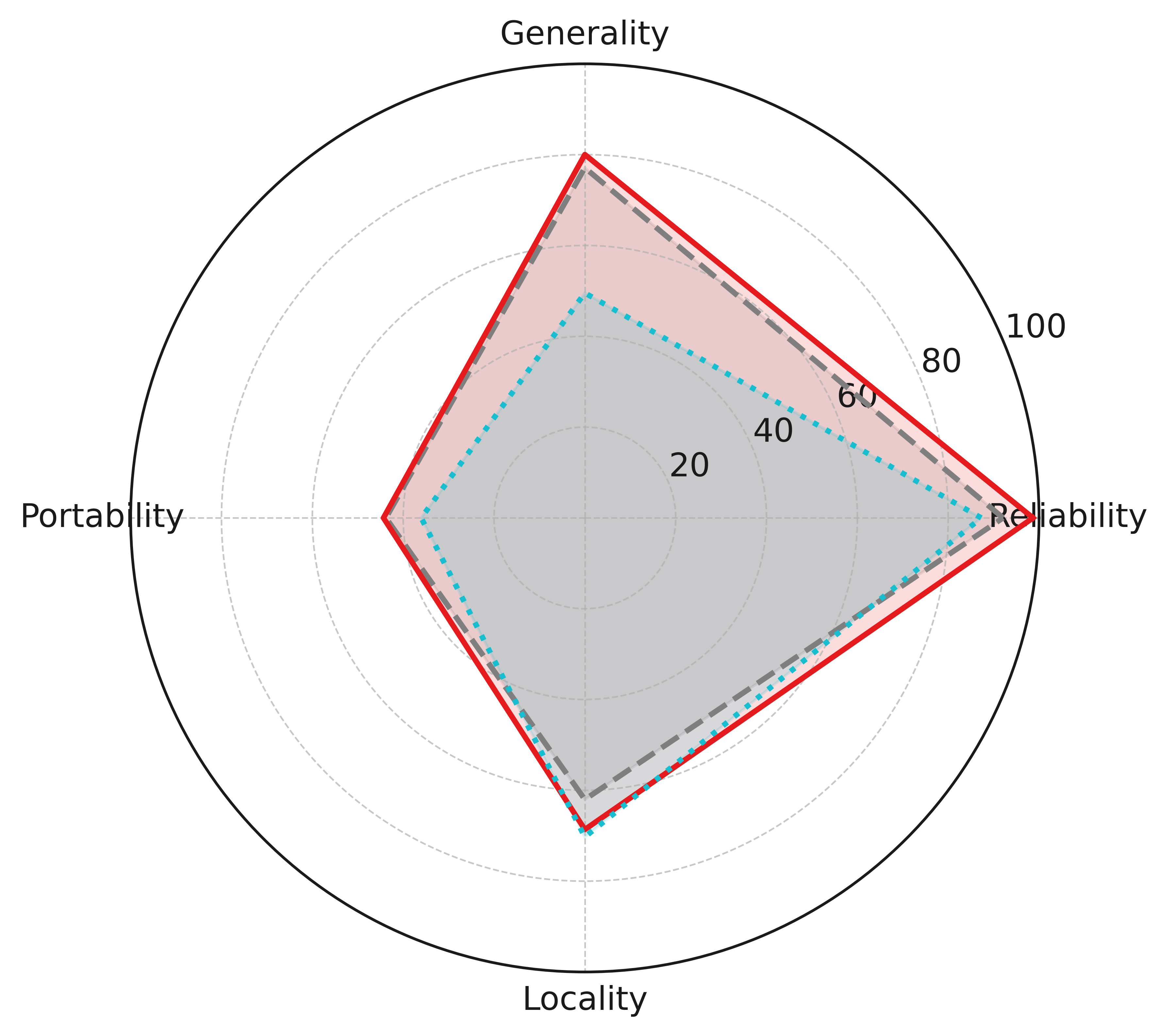}
         \caption{ROME}
         \label{fig:rome_languages_comparison_counterfact}
     \end{subfigure}
     \begin{subfigure}[b]{0.24\textwidth}
         \centering
         \includegraphics[width=\textwidth]{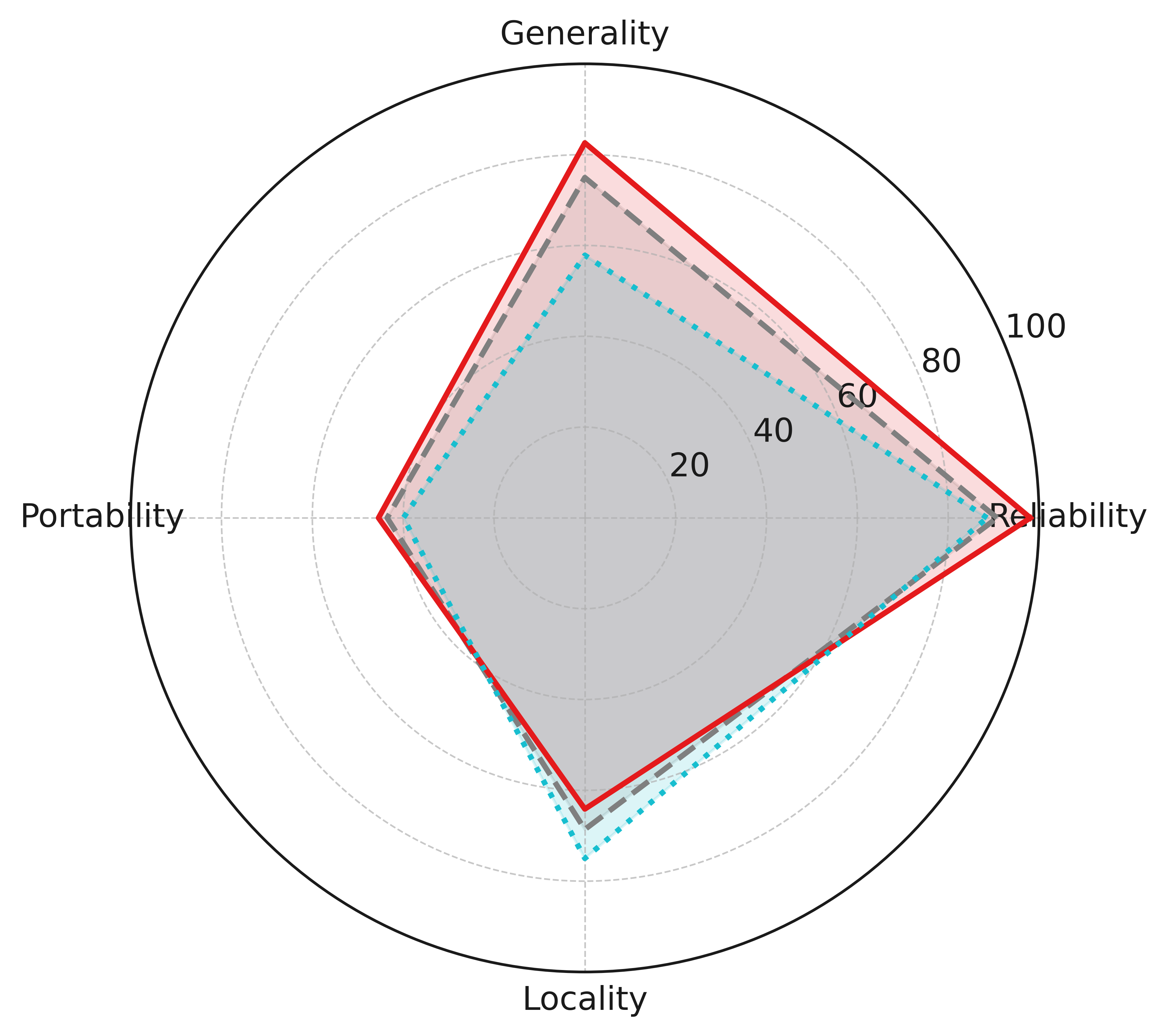}
         \caption{MEMIT}
         \label{fig:memit_languages_comparison_counterfact}
     \end{subfigure}
      \begin{subfigure}[b]{0.24\textwidth}
         \centering
         \includegraphics[width=\textwidth]{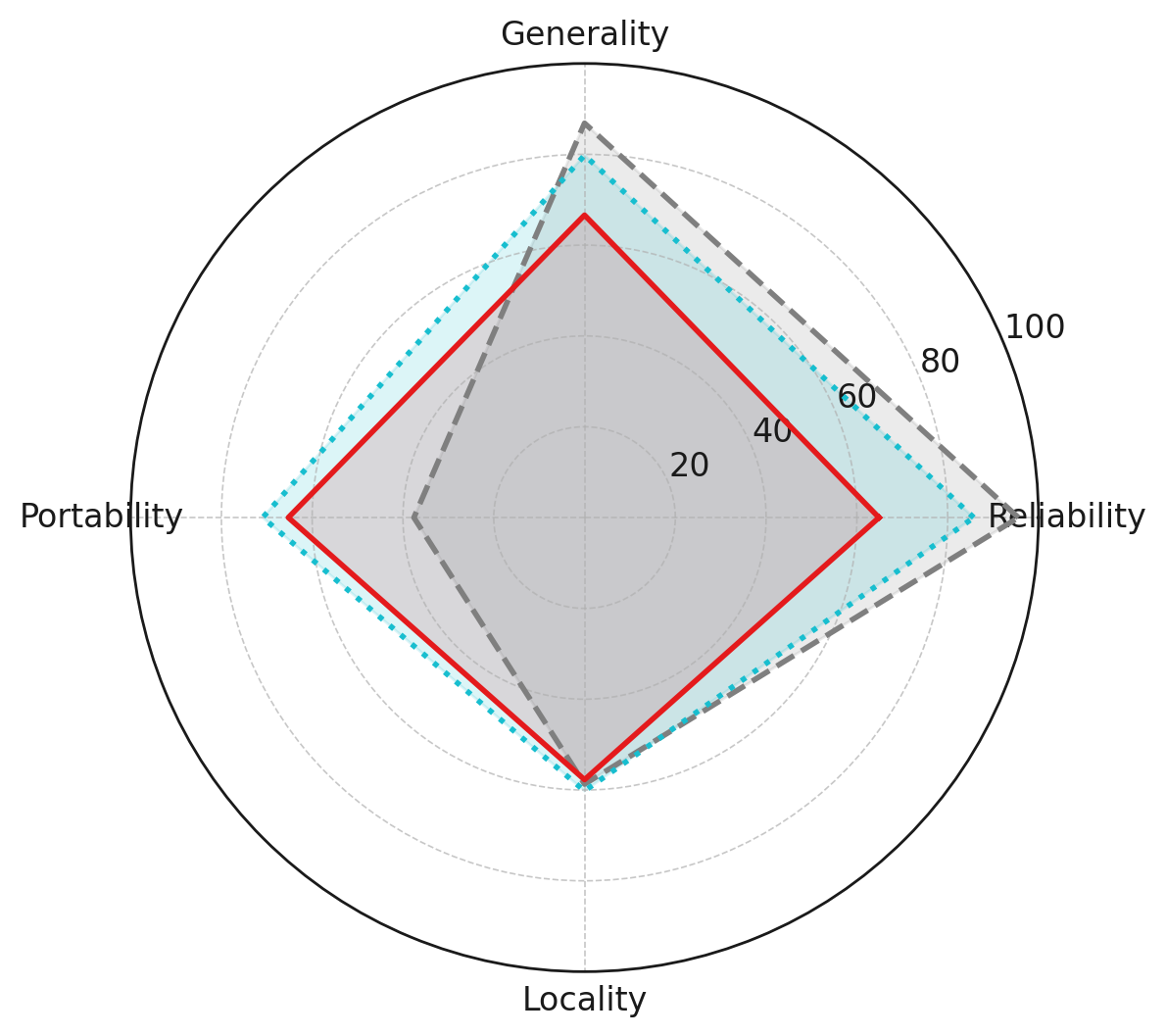}
         \caption{ICE}
         \label{fig:ike_languages_comparison_counterfact}
     \end{subfigure}
     \begin{subfigure}[b]{0.24\textwidth}
         \centering
         \includegraphics[width=\textwidth]{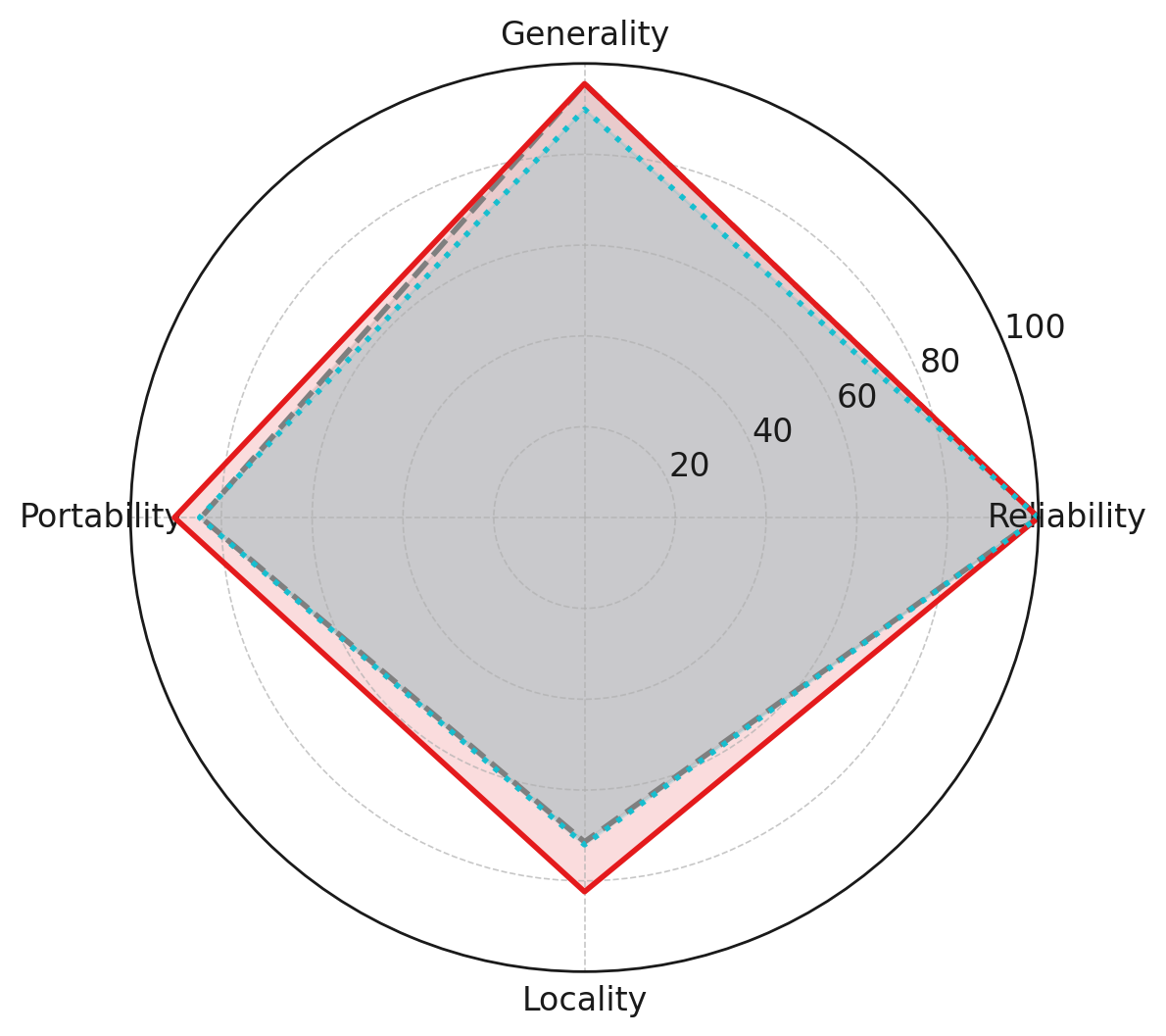}
         \caption{LTE}
     \end{subfigure}

    \caption{Impact of the editing language on the reliability, generality, portability and locality metrics on counterfact datasets for Llama2-7B-Chat}
    \label{fig:multilingual_comparison_counterfact}
\end{figure*}

\paragraph{Multilingual Comparison} The results of the multilingual comparison on the counterfact dataset are shown in figure \ref{fig:multilingual_comparison_counterfact}

\begin{figure*}
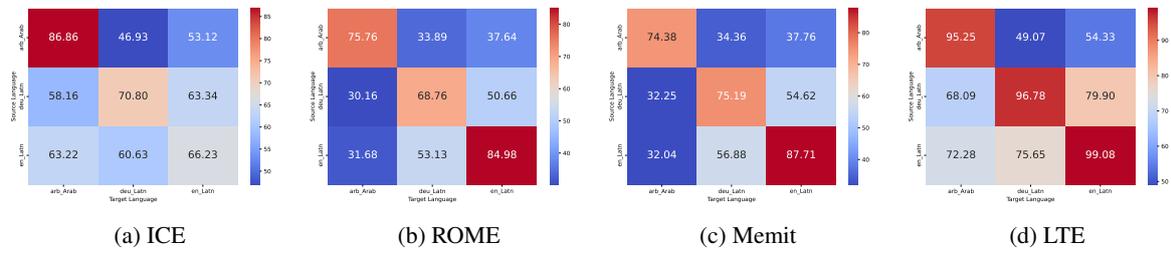

     \centering
     \begin{subfigure}[b]{0.24\textwidth}
         \centering
         \includegraphics[width=\textwidth]{figures/crosslingual-heatmaps-zsre/base_heatmap_rephrase_edit_success.png}
         \caption{ICE}
         \label{fig:ice_generality}
     \end{subfigure}
     \begin{subfigure}[b]{0.24\textwidth}
         \centering
         \includegraphics[width=\textwidth]{figures/crosslingual-heatmaps-zsre/rome-rephrase_edit_success.png}
         \caption{ROME}
         \label{fig:rome_generality}
     \end{subfigure}
     \begin{subfigure}[b]{0.24\textwidth}
         \centering
         \includegraphics[width=\textwidth]{figures/crosslingual-heatmaps-zsre/memit-rephrase_edit_success.png}
         \caption{Memit}
         \label{fig:memit_generality}
     \end{subfigure}
     \begin{subfigure}[b]{0.24\textwidth}
         \centering
         \includegraphics[width=\textwidth]{figures/crosslingual-heatmaps-zsre/english-tuned-final_heatmap_rephrase_edit_success.png}
         \caption{LTE}
         \label{fig:lte_generality}
     \end{subfigure}
    \caption{Cross Lingual Generality Metrics Comparison (ZsRE)}
    \label{fig:crosslingual_comparison_generality_zsre}
\end{figure*}

\begin{figure*}
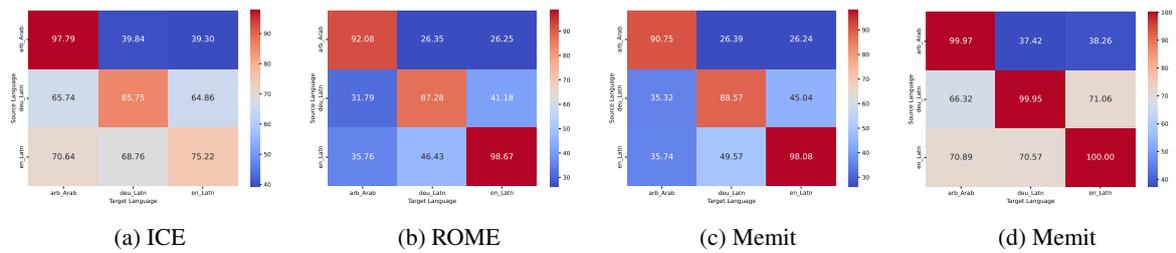

     \centering
     \begin{subfigure}[b]{0.24\textwidth}
         \centering
         \includegraphics[width=\textwidth]{figures/crosslingual-heatmaps-counterfact/base_heatmap_edit_success.png}
         \caption{ICE}
         \label{fig:ice_reliability_cf}
     \end{subfigure}
     \begin{subfigure}[b]{0.24\textwidth}
         \centering
         \includegraphics[width=\textwidth]{figures/crosslingual-heatmaps-counterfact/rome-edit_success.png}
         \caption{ROME}
         \label{fig:rome_reliaiblity_cf}
     \end{subfigure}
     \begin{subfigure}[b]{0.24\textwidth}
         \centering
         \includegraphics[width=\textwidth]{figures/crosslingual-heatmaps-counterfact/memit-edit_success.png}
         \caption{Memit}
         \label{fig:memit_reliability_cf}
     \end{subfigure}
     \begin{subfigure}[b]{0.24\textwidth}
         \centering
         \includegraphics[width=\textwidth]{figures/crosslingual-heatmaps-counterfact/english-tuned-final_heatmap_edit_success.png}
         \caption{Memit}
         \label{fig:lte_reliability_cf}
     \end{subfigure}
 
    \caption{Cross Lingual Reliability Metrics Comparison (Counterfact) }
    \label{fig:crosslingual_comparison_reliability_cf}
\end{figure*}

\begin{figure*}
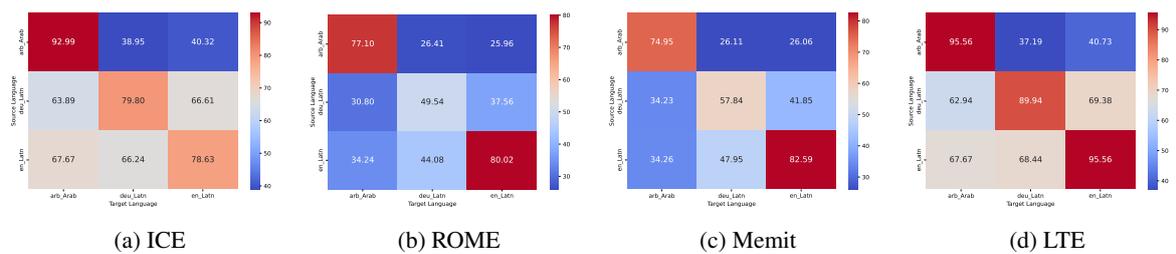

     \centering
     \begin{subfigure}[b]{0.24\textwidth}
         \centering
         \includegraphics[width=\textwidth]{figures/crosslingual-heatmaps-counterfact/base_heatmap_rephrase_edit_success.png}
         \caption{ICE}
         \label{fig:ice_generality_cf}
     \end{subfigure}
     \begin{subfigure}[b]{0.24\textwidth}
         \centering
         \includegraphics[width=\textwidth]{figures/crosslingual-heatmaps-counterfact/rome-rephrase_edit_success.png}
         \caption{ROME}
         \label{fig:rome_generality_cf}
     \end{subfigure}
     \begin{subfigure}[b]{0.24\textwidth}
         \centering
         \includegraphics[width=\textwidth]{figures/crosslingual-heatmaps-counterfact/memit-rephrase_edit_success.png}
         \caption{Memit}
         \label{fig:memit_generality_cf}
     \end{subfigure}
     \begin{subfigure}[b]{0.24\textwidth}
         \centering
         \includegraphics[width=\textwidth]{figures/crosslingual-heatmaps-counterfact/english-tuned-final_heatmap_rephrase_edit_success.png}
         \caption{LTE}
         \label{fig:lte_generality_cf}
     \end{subfigure}
 
    \caption{Cross Lingual Generality Metrics Comparison (Counterfact) }
        \label{fig:crosslingual_generality_cf}
\end{figure*}

\paragraph{Additional Cross-Lingual Results} The crosslingual generality metric on the ZsRE are shown in figure \ref{fig:crosslingual_comparison_generality_zsre} and the cross-lingual reliability and generality metric on counterfact are shown in figures \ref{fig:crosslingual_comparison_reliability_cf} \& \ref{fig:crosslingual_generality_cf}

\end{document}